\def\year{2021}\relax
\documentclass[letterpaper]{article} 
\usepackage{aaai21}  
\usepackage{times}  
\usepackage{helvet} 
\usepackage{courier}  
\usepackage[hyphens]{url}  
\usepackage{graphicx} 
\usepackage{natbib}  
\usepackage{caption} 
\usepackage[utf8]{inputenc} 
\usepackage[T1]{fontenc}    
\usepackage{hyperref}       
\usepackage{url}            
\usepackage{booktabs}       
\usepackage{amsfonts}       
\usepackage{nicefrac}       
\usepackage{microtype}      
\usepackage{comment}
\usepackage{amsmath,amssymb} 
\usepackage{color}
\usepackage{multirow}
\usepackage[linesnumbered,ruled]{algorithm2e}
\nocopyright
\title{Fitting the Search Space of Weight-sharing NAS\\with Graph Convolutional Networks}
\author{
    Xin Chen\textsuperscript{1},\quad
    Lingxi Xie\textsuperscript{1},\quad
    Jun Wu\textsuperscript{2},\quad
    Longhui Wei\textsuperscript{1},\quad
    Yuhui Xu\textsuperscript{3},\quad
    Qi Tian\textsuperscript{1}\\
}
\affiliations{

    \textsuperscript{1}Huawei Cloud \& AI,\qquad
    \textsuperscript{2}Fudan University,\qquad
    \textsuperscript{3}Shanghai Jiao Tong University\\
    {\small\texttt{chenxin061@gmail.com}}\qquad {\small\texttt{198808xc@gmail.com}}\qquad {\small\texttt{wujun@fudan.edu.cn}}\\
    {\small\texttt{weilonghui1@huawei.com}}\qquad {\small\texttt{yuhuixu@sjtu.edu.cn}}\qquad {\small\texttt{tian.qi1@huawei.com}}

}

\begin{document}

\maketitle

\newcommand{\revisionCR}{black}

\begin{abstract}
    Neural architecture search has attracted wide attentions in both academia and industry. To accelerate it, researchers proposed weight-sharing methods which first train a super-network to reuse computation among different operators, from which exponentially many sub-networks can be sampled and efficiently evaluated. These methods enjoy great advantages in terms of computational costs, but the sampled sub-networks are not guaranteed to be estimated precisely unless an individual training process is taken. This paper attributes such inaccuracy to the inevitable mismatch between assembled network layers, so that there is a random error term added to each estimation. We alleviate this issue by training a graph convolutional network to fit the performance of sampled sub-networks so that the impact of random errors becomes minimal. With this strategy, we achieve a higher rank correlation coefficient in the selected set of candidates, which consequently leads to better performance of the final architecture. In addition, our approach also enjoys the flexibility of being used under different hardware constraints, since the graph convolutional network has provided an efficient lookup table of the performance of architectures in the entire search space. 
\end{abstract}

Neural architecture search (NAS) is an emerging research field of automated machine learning (AutoML), with the goal being exploring deep networks that have not been investigated by manual designs. Early NAS approaches~\cite{zoph2016neural,real2017large} mostly sampled architectures from a large search space and evaluated them using an individual training-from-scratch process. Despite their ability in finding powerful architectures, the search process is often computationally expensive, \textit{e.g.}, hundreds or even thousands of GPU-days are required even if efficient sampling strategies were used~\cite{liu2018progressive,real2018regularized}.

To alleviate the computational burden, researchers started to consider reusing computation among differently sampled architectures~\cite{cai2018efficient}. Going one step forward, an efficient framework named weight-sharing NAS~\cite{pham2018efficient,liu2018darts,xie2020weight} was proposed in which the search space is formulated into a \textbf{super-network}, an over-parameterized architecture which contains the parameters of all architectures (often referred to as \textbf{sub-networks}) that can appear. With the super-network being pre-trained, each sampled sub-network can be estimated with reduced costs, so that the overall search process is largely accelerated, \textit{e.g.}, by $4$--$5$ orders of magnitudes if differentiable search is used~\cite{chen2019progressive,xu2020pc}.

One of the most important concerns of weight-sharing NAS lies in the accuracy of sub-network sampling. In other words, there is no guarantee if the estimated accuracy of the sub-networks can reflect the real performance of the corresponding architectures. In particular, given a search space, $\mathcal{S}$, and two architectures, $\mathbb{M}_1$ and $\mathbb{M}_2$, we can either evaluate them by training them from scratch or train a super-network and then sample them from it. Then, how often will the relative performance (\textit{i.e.}, whether $\mathbb{M}_1$ is better than $\mathbb{M}_2$) be consistent under the two evaluation methods? We perform experiments by sampling $8$ architectures with similar hardware complexity from a search space containing $19$ cells, each of which has $6$ possibilities. The result is disappointing: the Kendall-$\tau$ coefficient between two rankings is merely $0.2143$ ($0$ being random permutation). With such a low correlation, it is hard to guarantee that weight-sharing NAS can find a high-quality architecture in the search space.

To solve this problem, we make an assumption that the inaccuracy mostly comes from the randomness during training the super-network. We take a single-path training process~\cite{guo2019single,chu2019fairnas} as an example: in each training iteration, only one operator in each cell gets updated while others remain unchanged. For a candidate architecture being evaluated, its performance will potentially be high if it is closely related to one of the recently trained sub-networks. Therefore, for an arbitrary architecture, $\mathbb{M}$, we formulate the relationship between the real performance, $z_\mathbb{M}^\star$, and the estimated one, $z_\mathbb{M}$, into a linear formula\footnote{\textcolor{\revisionCR}{A higher-order function may better fit the data distribution, but it also increases the risk of over-fitting. More importantly, we found no cues in NAS that suggests a higher-order relationship between the true and sampled accuracy, so using a linear assumption is a relatively safe choice. }} of ${z_\mathbb{M}}={a\times z_\mathbb{M}^\star+\epsilon_\mathbb{M}+b}$, where $a$ and $b$ are constants and $\epsilon_\mathbb{M}$ is a zero-mean random variable associated to $\mathbb{M}$. The goal is to alleviate the impact of $\epsilon_\mathbb{M}$, since the remaining part will not change the relative ranking of the sampled architectures.

Next, we assume that $z_\mathbb{M}^\star$, though difficult (or expensive) to obtain, is a learnable function with respect to $\mathbb{M}$. Hence, we sample a set of training data, $\left\{\left(\mathbb{M},z_\mathbb{M}\right)\right\}$, and train a graph convolutional network $f\!\left(\cdot\right)$ by minimizing the average error of $\left|z_\mathbb{M}-f\!\left(\mathbb{M}\right)\right|$. We expect that $f\!\left(\cdot\right)$, by seeing large amount of training pairs, $\left(\mathbb{M},z_\mathbb{M}\right)$, can minimize of the random noise, \textit{i.e.}, ${f\!\left(\mathbb{M}\right)}\approx{a\times z_\mathbb{M}^\star+b}$. Hence, $f\!\left(\mathbb{M}\right)$ serves as an estimation of the performance of $\mathbb{M}$ which is almost free to compute ($f\!\left(\cdot\right)$ is very light-weighted) meanwhile being more reliable than $z_\mathbb{M}$ since it produces the same ranking as $z_\mathbb{M}^\star$.

We perform experiments on the search space defined by FairNAS~\cite{chu2019fairnas} which has $19$ cells and $6$ choices for each cell. Starting with the same super-network, our approach partitions the search space into three subspaces (with $7$, $6$, and $6$ cells, respectively) due to the limitation of CPU memory (it is difficult to support more than $6^7$ nodes in a graph). The iterative optimization takes $9$ GPU-hours beyond super-network training which is faster than that of FairNAS which sampled $12.8\mathrm{K}$ sub-networks directly. Our approach achieves a top-$1$ accuracy of $75.5\%$ on ImageNet with $383\mathrm{M}$ multi-add operations, or $75.6\%$ with $393\mathrm{M}$ multi-add operations, which consistently surpass the counterpart that uses random sampling. In addition, the graph convolutional network makes it easier to formulate the network performance in the entire search space into a lookup table, based on which a wide range of hardware constraints can be integrated towards more demands of architecture search.

\section{Related Work}

Neural architecture search aims to automate the network design process and discover architectures that perform better than hand-crafted ones~\cite{elsken2019neural}. In a general NAS pipeline, architectures are sampled from a pre-defined search space and evaluated on a specific task, \textit{e.g.}, image classification. To reduce the number of sampled architectures and accelerate the search process, heuristic algorithms including evolutionary algorithms (EA) and reinforcement learning (RL) are adopted to guide the sampling process. Recently, some EA-based~\cite{xie2017genetic,zoph2016neural,real2018regularized} and RL-based~\cite{zoph2018learning,liu2018progressive,tan2019mnasnet} approaches have achieved state-of-the-art performance on a variety of computer vision and natural language processing tasks. However, these approaches share a common evaluation scheme that optimizes each sampled architecture from scratch, which results in a critical drawback of heavy computational overhead even on CIFAR10, a small proxy dataset.

To alleviate the computational burden, researchers proposed an efficient solution which trains an over-parameterized super-network to cover all architectures in the search space and reuses or shares network weights among multiple architectures. This was named one-shot NAS~\cite{brock2017smash} or weight-sharing NAS~\cite{pham2018efficient}. 
SMASH~\cite{brock2017smash} proposed to train a HyperNet to generate weights for the evaluation of sampled architectures, reducing the search cost by $2$--$3$ orders of magnitudes.
ENAS~\cite{pham2018efficient} proposed to share weights among child models and apply reinforcement learning to improve the efficiency of computations, which dramatically reduce the search time to less than half a day with a single GPU. 
Pushing one-shot NAS to a continuous parameter space, DARTS~\cite{liu2018darts} and its variants~\cite{xu2020pc,chen2019progressive,dong2019search,xie2018snas,bi2020gold} adopted a differentiable framework that assigned a set of architectural parameters aside from the parameters of the super-network and iteratively optimized them by gradient descent, where the importance of different candidates is determined by the value of these architectural parameters. The reduction in computational burden facilitated architecture search on large-scale proxy datasets, \textit{e.g.}, ImageNet~\cite{deng2009imagenet}, with acceptable search cost.
ProxylessNAS~\cite{cai2018proxylessnas} searched architectures directly on ImageNet, where they proposed to train the one-shot super-network by sampling only one path each time with a binary mask and optimize architectural parameters pairwise. 
FBNet~\cite{wu2019fbnet} adopted a differentiable scheme that is similar to DARTS and searched for the optimal architecture on ImageNet in a chain-styled search space.

Despite the great success in accelerating NAS, one-shot or weight-sharing methods still suffer from a severe problem named ranking inconsistency, which refers to the fact that the estimation of a sub-network can be different when it is sampled from the super-network and when it is trained from scratch. Single-path one-shot NAS~\cite{guo2019single} used a uniformly sampling method to guarantee that all candidates are fully and equally trained, which is believed to be effective on alleviating the inconsistency. FairNAS~\cite{chu2019fast} paved one step further and proposed to train the super-network with a strict constraint on fairness and demonstrated a stronger correlation between the one-shot and stand-alone evaluation results.

\section{Our Approach}

\textcolor{\revisionCR}{Our work is based on the pipeline of weight-sharing NAS. Under this pipeline, $L\times O$ sets of parameters are trained to simulate the behavior of $O^L$ architectures, where each architecture consists of $L$ layers and each layer is chosen from $O$ candidates. Although this scheme demonstrates superior efficiency, there is no guarantee that the performance of sub-networks can be accurately estimated in this way. We will delve deep into this problem in the following parts. More details about weight-sharing NAS for classification can be found in the Appendix.}

\begin{figure*}[!t]
\begin{minipage}{0.49\linewidth}
    \centering
    \begin{tabular}{|l|c|ccc|}
    \hline
    ID & FLOPs & $\mathrm{Acc}^\star$ & $\mathrm{Acc}_{150}$ & $\mathrm{Acc}_{160}$ \\
    \hline
    01 & $400\mathrm{M}$ & $85.78\%$ & $81.59\%$ & $81.20\%$ \\
    02 & $401\mathrm{M}$ & $85.76\%$ & $81.56\%$ & $81.41\%$ \\
    03 & $401\mathrm{M}$ & $85.59\%$ & $81.73\%$ & $81.55\%$ \\
    04 & $400\mathrm{M}$ & $85.48\%$ & $81.95\%$ & $81.67\%$ \\
    05 & $403\mathrm{M}$ & $85.32\%$ & $81.70\%$ & $81.37\%$ \\
    06 & $399\mathrm{M}$ & $85.28\%$ & $81.64\%$ & $81.15\%$ \\
    07 & $402\mathrm{M}$ & $84.98\%$ & $81.60\%$ & $81.58\%$ \\
    08 & $400\mathrm{M}$ & $84.60\%$ & $81.53\%$ & $81.46\%$ \\
    \hline
    $\tau$ & N/A & $-$ & $0.2143$ & $-0.1429$ \\
    \hline
    \end{tabular}
\end{minipage}
\hfill
\begin{minipage}{0.49\linewidth}
    \centering
    \includegraphics[width=1.0\linewidth]{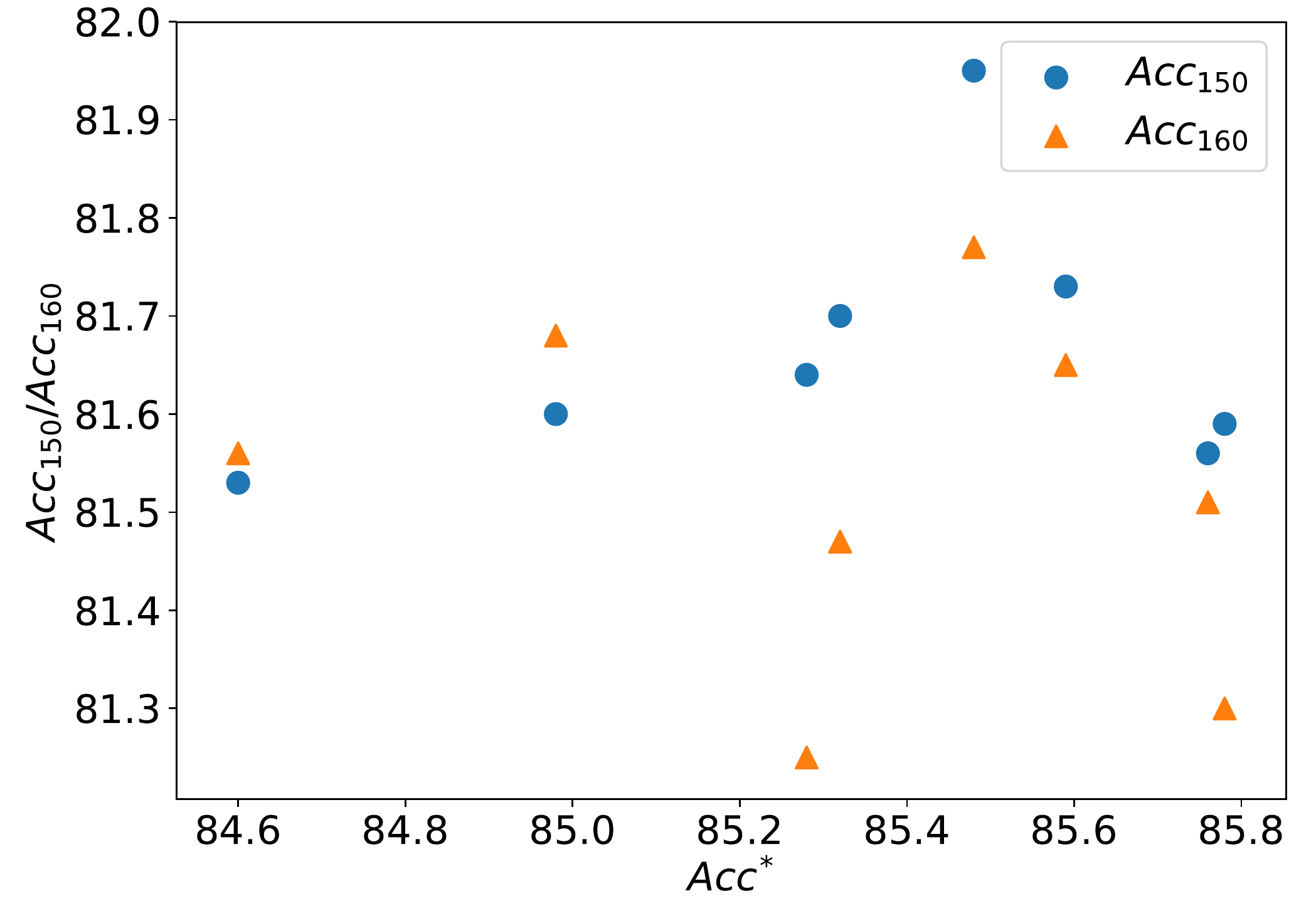}
\end{minipage}
\caption{The inconsistency between the performance of sub-networks by sampled from the super-network and trained from scratch. We choose $8$ architectures with similar FLOPs and use the super-networks after $150$ and $160$ epochs of training. \textbf{Left}: the table summarizing all classification accuracy over $100$ classes, in which $\tau$ indicate the Kendall-$\tau$ coefficient between the ranking and the ground-truth. \textbf{Right}: visualizing the results on a 2D plane, which the horizontal and vertical axes denote the ground-truth and sampled accuracy, respectively.}
\label{fig:inconsistency}
\end{figure*}

\subsection{Inaccuracy of Sub-network Sampling}

Throughout this paper, we work on the search space defined by FairNAS~\cite{chu2019fairnas} which has $19$ inverted residual cells~\cite{sandler2018mobilenetv2}, each of which has $6$ options differing from each other in the expansion ratio and kernel size. On the ImageNet dataset with $1\rm{,}000$ classes, we randomly sample $100$ classes and use the corresponding subset to optimize the super-network. $90\%$ of the training data is used to update the model parameters, and the remaining $10\%$ is used for evaluating each of the sampled sub-networks. Here, we follow~\cite{guo2019single,chu2019fairnas} to directly feed each testing image into the sampled sub-network and obtain the classification accuracy. Evaluating each sub-network on the validation subset (around $13\mathrm{K}$ images) takes an average of $4.48$ seconds on an NVIDIA Tesla-V100 GPU.

The first and foremost observation is that the performance of sub-networks cannot be accurately evaluated in this manner. To show this, we randomly sample $8$ architectures with a similar complexity (\textit{i.e.}, $400\mathrm{M}$ multi-adds), and evaluate their performance in two different ways, namely, sampling the corresponding sub-network from the well-trained super-network, or training the sub-network from scratch. Results are summarized in Figure~\ref{fig:inconsistency}. One can see that the ranking of the sampled accuracy can be very different from that of the ground-truth accuracy, \textit{i.e.}, when each architecture undergoes a complete training process from scratch. That being said, the architecture with the highest sub-network sampling accuracy may not be the optimal solution. As a side note, FairNAS~\cite{chu2019fairnas} advocated for that sub-network sampling is good at performance estimation, but we point out that the seemingly accurate prediction was mostly due to the large difference between the sampled architectures (they sampled $13$ architectures with FLOPs varying significantly, unlike the $8$ architectures sampled by our work that are all close to $400\mathrm{M}$ FLOPs).
    
We explain this phenomenon by noting that the single-path strategy of super-network training is highly random. Let $\mathbb{M}_0$ be an architecture to be evaluated using sub-network sampling. Since each layer of $\mathbb{M}_0$ shares the training process with other operators, it is probable that each layer of $\mathbb{M}_0$ gets updated in different training iterations. In this situation, when $\mathbb{M}_0$ is sampled as a sub-network, its layers may not `cooperate' well with each other. In particular, if a layer does not get updated for a long time, its parameters are relatively `outdated' and thus may incur a low recognition accuracy of $\mathbb{M}_0$. On the other hand, if all layers of another architecture, $\mathbb{M}_1$, happen to be updated sufficiently in the last few iterations, the recognition accuracy of $\mathbb{M}_1$ is potentially high. Nevertheless, this does not mean that $\mathbb{M}_1$ is better than $\mathbb{M}_0$.

In this paper, we introduce a simple model to formulate the above randomness introduced by sub-network sampling. Let $z_\mathbb{M}^\star$ be the ground-truth accuracy of $\mathbb{M}$, \textit{e.g.}, when $\mathbb{M}$ is trained individually, and $z_\mathbb{M}$ be the accuracy obtained by sub-network sampling. Note that $z_\mathbb{M}$ is related to the super-network, $\mathbb{S}$, but we omit $\mathbb{S}$ for simplicity. Our key assumption is that $z_\mathbb{M}$ can be written in a linear function of $z_\mathbb{M}^\star$ added by a random perturbation:
\begin{equation}
{z_\mathbb{M}}={a\times z_\mathbb{M}^\star+b+\epsilon_\mathbb{M}}.
\end{equation}
Here, $a$ is a linear coefficient indicating the systematic error between $z_\mathbb{M}^\star$ and $z_\mathbb{M}$, and $b$ is a constant bias between $z_\mathbb{M}^\star$ and $z_\mathbb{M}$. Most often, $a$ is slightly smaller than $1.0$ because the super-network is not sufficiently optimized compared to a complete training process. $\epsilon_\mathbb{M}$ is a random variable associated to $\mathbb{M}$. According to the above analysis, it is mostly determined by how the layers of $\mathbb{M}$ are updated in the final iterations, so it is totally random and unavailable unless $z_\mathbb{M}^\star$ is tested. As shown in Figure~\ref{fig:inconsistency}, the intensity of $\epsilon_\mathbb{M}$ can be large -- for two architectures, $\mathbb{M}_1$ and $\mathbb{M}_2$, it is possible that the difference between $\epsilon_{\mathbb{M}_1}$ and $\epsilon_{\mathbb{M}_2}$ is even larger than $a\times\left|z_{\mathbb{M}_1}^\star-z_{\mathbb{M}_2}^\star\right|$, which may alter the relative ranking between the two models.

\subsection{Further Analysis on the Key Assumption}

\textcolor{\revisionCR}{We perform a toy experiment to demonstrate the influence of the random error term. We adopt a small search space that contains $6$ cells, each of which has $6$ choices, {\em i.e.}, a search space with $6^6 = 46\rm{,}656$ candidate architectures. The super-network is firstly trained for $150$ epochs to guarantee a sufficient convergence degree and we denote the parameters of this super-network as $\boldsymbol{\theta}_{150}$. $10\rm{,}000$ candidate architectures (about $20\%$ of the whole search space) are randomly sampled from the search space and evaluated with $\boldsymbol{\theta}_{150}$ on the stand-alone validating set by sub-network sampling. We mark the selected architectures and their corresponding evaluation accuracy as $\left\{\left(\mathbb{M}_m,z_{\mathbb{M}_m}\right)\right\}_{m=1}^{10000}$. After this process, the super-network is further trained for a few more iterations\footnote{Here, we train the super-network for $12$ iterations, {\em i.e.}, about one-tenth of a full epoch. } with the same training data, which results with a set of slightly modified parameters, denoted as $\boldsymbol{\theta}_{150}'$. A similar process is performed with $\boldsymbol{\theta}_{150}'$ to obtain a set of new evaluation results, $\left\{\left(\mathbb{M}_m,z_{\mathbb{M}_m}'\right)\right\}_{m=1}^{10000}$.}

\textcolor{\revisionCR}{After acquiring these two set of evaluation results, we calculate the Kendall-$\tau$ coefficient between them and the result is astonishing. The Kendall-$\tau$ coefficient is only $0.5470$, which implies a big change on the ranking of selected architectures. This phenomenon strongly support our assumption that the random error term does matter and it can even alter the ranking of two architectures. 
}
\subsection{Alleviating Noise with Graph Convolutional Networks}\label{gcn_nas}

In what follows, we train a model to estimate ${a\times z_\mathbb{M}^\star+b}\equiv{z_\mathbb{M}-\epsilon_\mathbb{M}}$, \textit{i.e.}, eliminating the noise, $\epsilon_\mathbb{M}$, from $z_\mathbb{M}$. Note that $\epsilon_\mathbb{M}$ is not predictable, but we assume that it is a zero-mean noise, \textit{i.e.}, ${\sum_{\mathbb{M}\in\mathcal{S}}\epsilon_\mathbb{M}}={0}$, otherwise we can adjust the value of the bias, $b$, to achieve this goal. Then, we collect a number of pairs, $\left(\mathbb{M}_m,z_{\mathbb{M}_m}\right)$, and train a function, $f\!\left(\mathbb{M};\boldsymbol{\eta}\right)$, to fit $z_\mathbb{M}$, where $\boldsymbol{\eta}$ are learnable parameters. Since $\mathbb{M}$ is a structural data (the encoded architecture), we naturally choose a graph convolutional network (GCN)~\cite{kipf2017semi} to be the form of $f\!\left(\cdot\right)$\footnote{\textcolor{\revisionCR}{We could replace the GCN with a MLP or an LSTM here, but previous literature has proved that GCN is one of the optimal choices~\cite{shi2020bridging}.}}. The objective is written as:
\begin{equation}
\label{eqn:fitting}
{\boldsymbol{\eta}^\star}={\arg\min_{\boldsymbol{\eta}}\mathbb{E}\!\left[\left|f\!\left(\mathbb{M};\boldsymbol{\eta}\right)-z_\mathbb{M}\right|_1\mid\mathbb{M}\in\mathcal{S}\right]}.
\end{equation}
Under the assumption that $\epsilon_\mathbb{M}$ is completely irrelevant to $\mathbb{M}$ (\textit{i.e.}, for any $\boldsymbol{\eta}$ and an arbitrary set of architectures, the correlation coefficient between the $f\!\left(\mathbb{M};\boldsymbol{\eta}\right)$ and $\epsilon_\mathbb{M}$ values is $0$), we can derive that the optimal solution of Eqn~\eqref{eqn:fitting} is ${f\!\left(\mathbb{M}\right)}\equiv{a\times z_\mathbb{M}^\star+b}\equiv{z_\mathbb{M}-\epsilon_\mathbb{M}}$. In other words, the best $\mathbb{M}$'s that maximize $f\!\left(\mathbb{M}\right)$ and $z_\mathbb{M}^\star$ are the same, \textit{i.e.}, the optimal architecture, $\mathbb{M}^\star$.

The overall pipeline of using GCN for NAS is illstrated in Algorithm~\ref{alg:pipeline}. Since the search space is very large, we cannot include all $6^{19}$ architectures in one graph, so we start with an initialized model, $\mathbb{M}_0$, and apply an iterative process, each round of which updates a subset of layers of the current architecture. The idea that gradually optimizes the architecture is similar to that explored in PNAS~\cite{liu2018progressive}. Thanks to the stable property of GCN, the choice of $\mathbb{M}_0$ barely impacts the final architecture. In this paper, we choose the `standard' network in which all cells have a kernel size of $3$ and an expansion ratio of $6$, \textit{i.e.}, the architecture is similar to MobileNet-v2~\cite{sandler2018mobilenetv2}.

Let there be $T$ rounds of iteration in the search process. The $t$-th round, ${t}={0,1,\ldots,T-1}$, allows a subset of layers, $\mathcal{L}_t$, to be searched, and keeps all others fixed. The optimal sub-architecture found in this process will replace the corresponding layers, advancing $\mathbb{M}_{t}$ to $\mathbb{M}_{t+1}$, and the iterative process continues. Denote the subset of architectures that can appear in the $t$-th round by ${\mathcal{S}_t}\subset{\mathcal{S}}$. We construct a graph, ${\mathcal{G}_t}={\left(\mathcal{V}_t,\mathcal{E}_t\right)}$, where $\mathcal{V}_t$ denotes the set of nodes (each node corresponds to a sub-network) and $\mathcal{E}_t$ denotes the set of edges connecting nodes. For simplicity, we use the most straightforward way of graph construction, \textit{i.e.}, ${\mathcal{V}_t}\equiv{\mathcal{S}_t}$, and there exists a connection between two nodes if and only if the corresponding sub-networks have a Hamming distance of $1$. We observe in experiments that introducing more edges (\textit{e.g.}, using a Hamming threshold of $2$) does not improve GCN prediction accuracy but considerably increases training complexity.

Note that our work is related to prior ones focusing on predicting network performance without actually training them from scratch~\cite{luo2018neural,friede2019variational}. These approaches achieved promising results on some NAS benchmarks~\cite{ying2019bench,dong2020bench}, demonstrating that one can find top-ranked architectures by sampling a very small portion of the entire space. However, they lack the guarantee that the architectures in training, in particular those with very different topologies, have potential relationship to each other so that the trained model is indeed learning useful knowledge. Our work adds two assumptions that makes the prediction more reliable: (i) a well-trained super-network is provided so that the sampled sub-networks share the same set of network weights; (ii) only sub-networks with $1$-cell difference are related to each other. \textcolor{\revisionCR}{There are also other methods apply GCN to predictor-based NAS~\cite{shi2020bridging,wen2020neural,ning2020generic}. They mostly denotes one single architecture as a graph, while our method regards each single architecture as a vertex in the graph which represents the whole search space.}

Below we describe the details of our approach. Step 0 is performed only once, and Steps 1--3 can be iteratively executed until convergence or a pre-defined number of rounds is achieved.

\begin{algorithm}[!t]
\SetKwInOut{Input}{Input}
\SetKwInOut{Output}{Output}
\SetKwInOut{Return}{Return}
\Input{
Search space $\mathcal{S}$, dataset $\mathcal{D}$, cell index set $\mathcal{L}$;
}
\Output{
Optimal architecture $\mathbb{M}^\star$;
}
Split $\mathcal{D}$ into $\mathcal{D}_\mathrm{train}\cup\mathcal{D}_\mathrm{val}$, train the super-network $\mathbb{S}$ on $\mathcal{D}_\mathrm{train}$;\\
Initialize $\mathbb{M}_0$ as a default architecture, ${t}\leftarrow{0}$;\\
\Repeat{${t}={T}$ \textbf{or} $\mathrm{convergence}$}{
Sample a subset of active cells, $\mathcal{L}_t$, determine the subspace $\mathcal{S}_t$;\\
Construct a graph ${\mathcal{G}_t}={\left(\mathcal{V}_t,\mathcal{E}_t\right)}$, in which ${\mathcal{V}_t}\equiv{\mathcal{S}_t}$;\\
Sample $M$ architectures from $\mathcal{S}_t$ and evaluate the performance to fill up part of $\mathcal{G}_t$;\\
Train a GCN on $\mathcal{G}_t$, use the GCN to find top-$K$ architectures in $\mathcal{S}_t$;\\
Evaluate all $K$ architectures to find the best one, $\mathbb{M}_t^\star$;\\
${\mathbb{M}_{t+1}}\leftarrow{\mathbb{M}_t^\star}$, ${t}\leftarrow{t+1}$;
}
\Return{
$\mathbb{M}_t$.
}
\caption{
Applying GCN for Weight-sharing NAS
}
\label{alg:pipeline}
\end{algorithm}

\begin{figure*}[!t]
\centering
\includegraphics[width=0.90\linewidth]{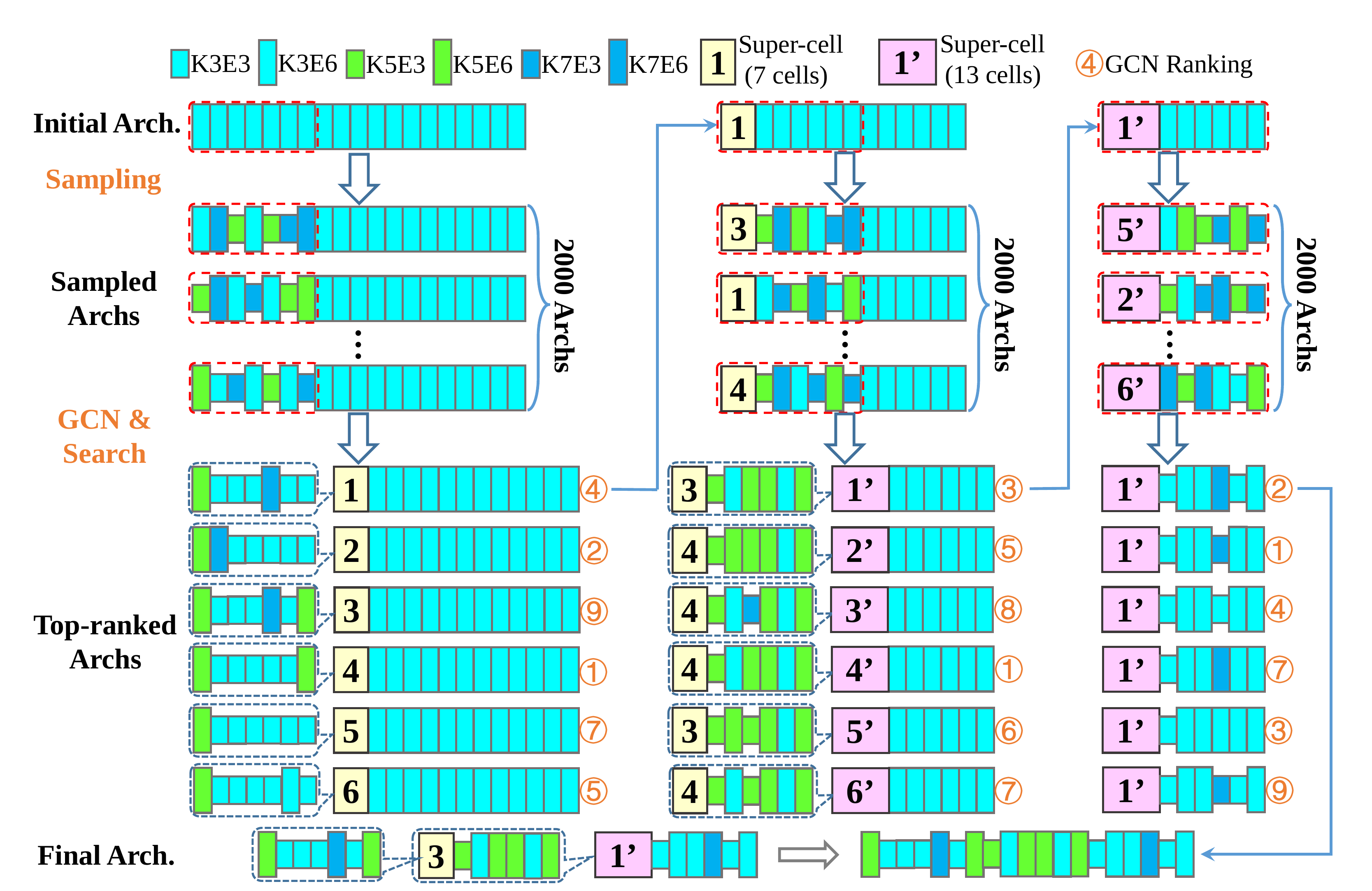}
\caption{A typical process of how GCN assists to find the best architecture (best viewed in color).  In the left, a segment of $7$ cells is firstly considered. Architectures are uniformly sampled from the subspace and evaluated with the super-network. Then these sampled architectures are encoded to train a GCN and performance of all architectures in the subspace can be predicted with the learned GCN. $6$ top-ranked sub-architectures (ranked by sampling top sub-networks predicted by GCN) are preserved and act as super-cells in the search process of the next segment. After all cells have been searched, we choose the top-1 architecture from the final segment and replace the super-cells with their corresponding sub-architectures to form the best network architecture.
}
\label{fig:process}
\end{figure*}

\noindent$\bullet$\quad\textbf{Step 0: Super-network Training}

We train the super-network in a single-path, one-shot process. During each iteration, a mini-batch is sampled from the training set, and a sub-network is constructed by randomly choosing a cell (building block) for each layer. Following FairNAS~\cite{chu2019fairnas}, the candidate cells in each layer are always sampled with equal probability, regardless of the current status and how these cells contribute to network accuracy. This strategy is believed to improve the `fairness' of sampling, \textit{i.e.}, in the probabilistic viewpoint, all sub-networks have the equal chance of being sampled and optimized. Thus, the relative ranking among sub-networks is believed to be more accurate.

The length of the training stage is not very important, as we will show in the experimental section that the fitting degree of the super-network does not heavily impact the searched architecture. $150$ epochs is often sufficient for super-network training, which takes around $13$ hours on eight GPUs. 

\noindent$\bullet$\quad\textbf{Step 1: Sub-network Sampling}

In the $t$-th round of iteration, given the set of layer indices to be searched, $\mathcal{L}_t$, we define a subspace of $\mathcal{S}$, denoted by $\mathcal{S}_t$. Let ${\mathcal{V}_t}={\mathcal{S}_t}$, from which a small subset of sub-networks are uniformly sampled and evaluated. Let the number of sampled architectures be $M$, and the collected architecture-performance pairs be $\left\{\left(\mathbb{M}_m,z_{\mathbb{M}_m}\right)\right\}_{m=1}^M$. Due to the potentially heavy computational overhead, $M$ is often much smaller than the total number of sub-networks, $\left|\mathcal{V}_t\right|$.


\noindent$\bullet$\quad\textbf{Step 2: Training the GCN}

One of the key factors of GCN is named the adjacent matrix, $\mathbf{A}$, which defines how two nodes (\textit{i.e.}, sub-networks) with an edge connection relate to each other and the way that we use the accuracy of one to predict that of another. For simplicity, we simply define $\mathbf{A}$ to be the similarity matrix, \textit{i.e.}, the weight of each edge is determined by the similarity between two nodes it connects. The last factor of GCN is the feature representation of each node (sub-network). We simply follow the Gray coding scheme to encode each sub-network into a fixed-dimensional vector. More details about the graph construction can be found in the Appendix.

We follow a standard training procedure described in~\cite{kipf2017semi} to train the GCN, yet we modify the classification head of the network (applied to node features) into a regression module to predict sub-network accuracy. After GCN training is complete, we can predict the accuracy of each sub-network with its feature vector and the similarity matrix, $\mathbf{A}$, with a negligible cost (evaluating all sub-networks in $\mathcal{S}_t$ takes only a few seconds in one GPU).

\noindent$\bullet$\quad\textbf{Step 3: Updating the Optimal Architecture}

After the optimization is finished, the GCN provides a lookup table that requires negligible costs in estimating the performance of each sub-network in the space of $\mathcal{S}_t$. As we elaborate previously, although GCN is good at depicting the property of the overall search space, its prediction in each single sub-network is not always accurate. So, we enumerate over the search space, find a number of sub-networks with the best performance, and perform sub-network evaluation again to determine the best of them. Experiments show that this re-verification step, with negligible costs, brings a consistent accuracy gain of $0.1\%$--$0.2\%$. Finally, the searched one is used to replace the corresponding layers of $\mathbb{M}_t$ and $\mathbb{M}_t$ becomes $\mathbb{M}_{t+1}$. If the search process does not terminate, then we go back to Step 1 and repeat data collection, GCN training, and architecture update.

It is possible that the optimal sub-network found in $\mathcal{S}_t$ does not lead to global optimality. To improve flexibility, we preserve $K$ top-ranked sub-networks and feed them into the next round. 
To achieve this goal, the searched layers in this ($t$-th) round are not allowed to be searched in the next ($t+1$-st) round, \textit{i.e.}, ${\mathcal{L}_t\cap\mathcal{L}_{t+1}}={\varnothing}$. Hence, the next round can sample each of the $K$ preserved sub-architectures as an entire -- in other words, the cells in $\mathcal{L}_t$ form a `super-cell' in the next round, and each preserved architecture is a choice in the super-cell. We will see in experiments that this strategy improves the stability of search, \textit{i.e.}, the chance of finding high-quality architectures is increased. 

\section{Experiments}

All experiments are conducted on ILSVRC2012~\cite{russakovsky2015imagenet}. Additional hyper-parameter settings and implementation details can be found in the Appendix. 

\subsection{A Typical Search Procedure: How GCN Works?}\label{subsection:typical}

Figure~\ref{fig:process} shows a complete search process. The search starts with an initial model, in which all cells have a kernel size of $3$ and an expansion ratio of $6$, denoted by \textsf{K3E6}. After searching in the first subspace containing $7$ cells, $6$ top-ranked sub-architectures are preserved and sent into the next stage as a super-cell. This process continues two more times until the final architecture is obtained.

From this example, we provide some intermediate statistics to show how GCN assists in the architecture search process. In the first subspace, $2\rm{,}000$ sub-networks are sampled, with the best one (denoted by $\tilde{\mathbb{M}}_0$) reporting an accuracy of $81.25\%$ (over the sampled $100$ classes). However, after GCN is trained and $100$ best sub-networks are chosen from its prediction, $\tilde{\mathbb{M}}_0$ does not appear in these $100$ sub-networks. Yet, when these $100$ sub-networks are sent into sub-network sampling, the best one ($\mathbb{M}_0^\star$) reports an accuracy of $81.11\%$ which is slightly lower than that of $\tilde{\mathbb{M}}_0$. But, this does not mean that $\tilde{\mathbb{M}}_0$ is indeed better -- it is the noise introduced by sub-network sampling that surpasses the difference between their actual performance. To verify this, we perform another NAS process that does not involve using GCN for prediction but simply preserves the top-ranked architectures from sub-network sampling. This strategy reports a final accuracy (over $1\rm{,}000$ classes) of $75.2\%$ (with 438M multi-adds), which is inferior to that ($75.5\%$) with GCN applied.

Two more evidences verify the effectiveness of GCN. On the one hand, GCN is able to fit the data collected by sub-network sampling. In three search segments, using $200$ left-out architecture-performance pairs for validation, the Kendall-$\tau$ coefficient (ranged in $\left[-1,1\right]$ with $0$ implying random guess) is always higher than $0.5$, \textit{i.e.}, the relative ranking of more than $75\%$ pairs is consistent between sub-network sampling and GCN prediction. On the other hand, GCN prediction actually works better than sub-network sampling. We choose $8$ architectures from the third subspace and evaluate their real performance by training them from scratch. The Kendall-$\tau$ coefficient between the real performance and sub-network sampling is $0.2143$ (still close to random guess), but the coefficient is improved to $0.6429$ by GCN. 

\textcolor{\revisionCR}{Furthermore, we perform linear regression on $\mathrm{Acc}_{150}$, $\mathrm{Acc}_{160}$ and the GCN-predicted one, $\mathrm{Acc}_{GCN}$, and re-plot corresponding dots and curves in the Appendix to show the results that GCN helps to improve the ranking consistency. The regression scores\footnote{The regression score is a criterion to measure the determination of prediction. Larger regression score is better. } of $\mathrm{Acc}_{150}^\mathrm{reg}$, $\mathrm{Acc}_{160}^\mathrm{reg}$ and $\mathrm{Acc}_\mathrm{GCN}^\mathrm{reg}$ are 0.07, 0.04 0.62, respectively. The large difference on regression scores between the super-network-evaluated accuracy and the GCN-predicted one is a proof that the random error term in our key assumption is largely suppressed. }


\begin{table*}[!t]
\begin{center}
\caption{Comparison with state-of-the-art architectures on ImageNet (mobile setting). $^\dagger$: SE module included, $^\ddagger$: estimated by~\cite{cai2018proxylessnas}, GB: gradient-based, BO: Bayesian Optimization. }
\label{table:sota}
\begin{tabular}{lcccccc}
\hline
\textbf{\multirow{2}{*}{Architecture}} & \multicolumn{2}{c}{\textbf{Test Acc. (\%)}} & \textbf{Params} & $\times+$ & \textbf{Search Cost} & \textbf{Search} \\
&                            \textbf{top-1} & \textbf{top-5} & \textbf{(M)} & \textbf{(M)} & \textbf{(GPU-days)} &  \textbf{Method}\\
\hline
Inception-v1~\cite{szegedy2015going}          & 69.8 & 89.9 & 6.6  & 1448 & -    & manual \\
MobileNet~\cite{howard2017mobilenets}         & 70.6 & 89.5 & 4.2  & 569  & -    & manual \\
ShuffleNet 2$\times$ (v1)~\cite{zhang2018shufflenet} & 73.6 & 89.8 & $\sim$5 & 524  & -    & manual \\
ShuffleNet 2$\times$ (v2)~\cite{ma2018shufflenet}    & 74.9 & -    & $\sim$5 & 591  & -    & manual \\
\hline
NASNet-A~\cite{zoph2018learning}              & 74.0 & 91.6 & 5.3 & 564 & 1800 & RL \\
NASNet-B~\cite{zoph2018learning}              & 72.8 & 91.3 & 5.3 & 488 & 1800 & RL \\
NASNet-C~\cite{zoph2018learning}              & 72.5 & 91.0 & 4.9 & 558 & 1800 & RL \\
AmoebaNet-A~\cite{real2018regularized}        & 74.5 & 92.0 & 5.1 & 555 & 3150 & EA \\
AmoebaNet-B~\cite{real2018regularized}        & 74.0 & 91.5 & 5.3 & 555 & 3150 & EA \\
AmoebaNet-C~\cite{real2018regularized}        & 75.7 & 92.4 & 6.4 & 570 & 3150 & EA \\
PNAS~\cite{liu2018progressive}                & 74.2 & 91.9 & 5.1 & 588 & 225  & SMBO \\
DARTS (second order)~\cite{liu2018darts}      & 73.3 & 91.3 & 4.7 & 574 & 4.0  & GB \\
SNAS (mild constraint)~\cite{xie2018snas}     & 72.7 & 90.8 & 4.3 & 522 & 1.5  & GB \\
P-DARTS (CIFAR10)~\cite{chen2019progressive}  & 75.6 & 92.6 & 4.9 & 557 & 0.3  & GB \\
PC-DARTS (ImageNet)~\cite{xu2020pc}           & 75.8 & 92.7 & 5.3 & 597 & 3.8  & GB \\
BONAS~\cite{shi2020bridging}                  & 75.4 & 92.5 & 5.1 & 557 & 7.5  & BO \\
\hline
MnasNet-92~\cite{tan2019mnasnet}              & 74.8 & 92.0 & 4.4 & 388 & 1667$^\ddagger$ & RL \\
MnasNet-A3$^\dagger$~\cite{tan2019mnasnet}    & 76.7 & 93.3 & 5.2 & 403 & 1667$^\ddagger$ & RL \\
ProxylessNAS (GPU)~\cite{cai2018proxylessnas} & 75.1 & 92.5 & 7.1 & 465 & 8.3 & GB \\
FBNet-C~\cite{wu2019fbnet}                    & 74.9 &  -   & 5.5 & 375 & 9.0 & GB \\
DenseNAS-C~\cite{fang2019densely}             & 74.2 & 91.8 & 6.7 & 383 & 3.8 & GB \\
FairNAS-A~\cite{chu2019fairnas}               & 75.3  & 92.4 & 4.6 & 388 & 12  & EA \\
\hline
One-Shot-NAS-GCN (ours)                       & 75.5 & 92.7 & 4.4 & 383 & 4.7 & GCN \\
One-Shot-NAS-GCN$^\dagger$ (ours)             & 76.6 & 93.1 & 4.6 & 384 & 4.7 & GCN \\
\hline
\end{tabular}
\end{center}
\end{table*}

\subsection{Comparison to the State-of-the-Arts}

The performance comparison on ImageNet classification with state-of-the-art architectures is listed in Table~\ref{table:sota}. For a fair comparison, we do not list architectures that are either trained with additional data augmentations (\textit{e.g.}, AutoAugment~\cite{cubuk2019autoaugment}) or equipped with extra architecture modifications (\textit{e.g.}, MobileNet-v3~\cite{howard2019searching} introduced H-Swish and reported a top-1 accuracy of 75.2\%). With a comparable amount of multi-add operations, the performance of our approach is on par with state-of-the-art methods that searched on a chain-styled search space. Our searched architecture enjoys a superior test accuracy than FairNAS~\cite{chu2019fairnas} due to a better fitting to the search space, which is explained in Section~\ref{subsection:typical}. Compared to those architectures searched on the DARTS-based search space, our discovered architecture achieves comparable performance while enjoys a smaller model size and multi-adds count. Notably, with nearly $20$M less multi-adds, the performance of our searched architecture is comparable to that of MnasNet-A3, which was searched by sampling and evaluating architectures by training-from-scratch and cost thousands of GPU-days. 

\section{Diagnostic Studies}\label{diagnosis}

\subsection{Relationship to the Overall Quality of the Super-network}

The length of one-shot training is worth discussion. For example, training the super-network throughout $150$ epochs achieves a $86.17\%$ training accuracy, and increasing the number of epochs to $450$ improves the accuracy to $94.67\%$, which potentially leads to a higher quality of sub-networks. However, such improvement does not necessarily cause the final architecture to be better, implying that the architecture-related random noise, $\epsilon_\mathbb{M}$, is still a major concern.

\begin{table}[h]
\setlength{\tabcolsep}{0.2cm}
\begin{center}
\caption{Results for different number of architectures sampled for each round and similarity type. $\tau_t$ denotes the Kendall-$\tau$ coefficient of the $t$-th round. A: Assigned; M: Measured. }
\label{table:ablation}
\begin{tabular}{cc|ccc|c}
\hline
\# Archs & Sim. & $\tau_1$ &  $\tau_2$ & $\tau_3$ & Test Acc. (\%)\\
\hline
2,000  & A & 0.76 & 0.58 & 0.51 & 75.48\\
2,000  & M & 0.74 & 0.60 & 0.52 & 75.47\\
5,000  & A & 0.74 & 0.60 & 0.50 & 75.47\\
5,000  & M & 0.75 & 0.62 & 0.53 & 75.45\\
\hline
\end{tabular}
\end{center}
\end{table}

\subsection{The Number of Sampled Architectures}

The number of architectures sampled for each round is a critical factor related to the search cost since the graph construction and GCN training only take a few minutes. We have tested two different settings of $M=2\rm{,}000$ and $M=5\rm{,}000$. The results are listed in Table~\ref{table:ablation} and no obvious difference is observed on both Kendall-$\tau$ coefficients and the evaluation accuracy of the discovered architectures. It takes around $9$ hours on a single GPU to finish the search process when $M=2\rm{,}000$, while the search cost is increased to about $21$ GPU-hours when we use the $M=5\rm{,}000$ setting. Thus, we adopt the former setting for most of the experiments. 

\subsection{The Similarity Measurement in GCN}

We have compared the influence of the way to define the similarity matrix and demonstrate the results in Table~\ref{table:ablation}. Due to the fact that both types of similarity definitions involve approximation, there is no obvious difference between them. 

\subsection{Preserving 1 vs. 6 Candidates}

Preserving multiple candidates from the previous round is important because more candidates can increase the probability that the most promising sub-architectures are included in the final searched architecture. We conduct experiments to validate it. We run the search process for $3$ times for both the settings of preserving $1$ candidate and $6$ candidates from the previous round. The classification accuracy of the former setting is $75.34\%\pm0.08\%$, while the latter is $75.47\%\pm0.01\%$, which supports our previous analysis. 

\subsection{Flexible NAS with Hardware Constraints}\label{flex_nas}

Thanks to the prediction ability of the trained GCN, we can easily build a lookup table that contains the predicted performance of all sub-networks in the search space (or subspace), upon which architectures that satisfy different hardware constraints can be selected. Without such a lookup table, the target architecture must be included in the explored architectures, which is rigid and inconvenient when multiple architectures with different constraints are required. We pick multiple architectures with different multi-adds levels by filtering out the top-ranked architectures that meet the target constraint and selecting the top-$1$ according to the sampling and re-evaluation ranking. As detailed results shown in the Appendix~\ref{hardware}, the selected architectures achieve $75.6\%$, $75.5\%$ and $75.4\%$ top-1 accuracies with $393$M, $383$M and $360$M multi-adds, respectively. Under target hardware constraints, our searched architectures keep comparable performance with state-of-the-art architectures under similar constraints. In addition, one can freely generalize our approach to other types of hardware constraints, \textit{e.g.}, network latency.

\section{Conclusions}

This paper introduces a novel idea that uses graph convolutional network to assist weight-sharing neural architecture search. The most important opinion is that there is an inevitable random noise between a well-trained super-network and the sub-networks sampled from it, and GCN, by averaging over the entire search space, can eliminate the error systematically and avoid the architecture search from falling into local minima. Experiments demonstrate the effectiveness and efficiency of our approach, in particular in the scenarios with additional hardware constraints. 

Our research paves the way of projecting the architectures in the search space into another low-dimensional space. This is a new direction which may provide new insights to neural architecture search, but currently, there are still some important issues that remain uncovered. For example, it is well known that fine-tuning each sampled sub-network can improve the accuracy of estimation, but it requires considerable computation -- this is a tradeoff between accuracy and efficiency. We can perform fine-tuning on a small number of sub-networks and assign the remaining ones to be evaluated by sub-network evaluation. The property of GCN in such a heterogeneous graph is worth investigating, which, as well as other topic, will be left for future work.

\section*{Acknowledgement}
We thank Dr. Song Bai and Zhengsu Chen for their valuable suggestions.

\begin{quote}
\begin{small}
    \bibliography{egbib}
\end{small}
\end{quote}

\newpage
\newpage
\begin{center}
\begin{huge}
\appendix{\textbf{Supplementary}}
\end{huge}
\end{center}
\section{\textcolor{\revisionCR}{Weight-sharing NAS for Classification}}

Most existing NAS approaches start with a search space, $\mathcal{S}$, which contains a large number of network architectures, denoted by ${\mathbb{M}}\in{\mathcal{S}}$. There is a proxy dataset ${\mathcal{D}}={\left\{\left(\mathbf{x}_n,\mathbf{y}_n\right)\right\}_{n=1}^N}$ with $N$ training samples, where $\mathbf{x}_n$ is the input, $\mathbf{y}_n$ is the one-hot output vector, and $y_n$ is the class label. The goal is to find the optimal architecture, $\mathbb{M}^\star$, which potentially generalizes to unseen testing data. To find $\mathbb{M}^\star$, a common flowchart is to partition $\mathcal{D}$ into training and validation subsets, ${\mathcal{D}}={\mathcal{D}_\mathrm{train}\cup\mathcal{D}_\mathrm{val}}$, use $\mathcal{D}_\mathrm{train}$ to optimize model parameters (\textit{e.g.}, convolutional weights), and use $\mathcal{D}_\mathrm{val}$ to validate if the trained model works well. Mathematically, this involves solving the following optimization problem:
\begin{tiny}
\begin{eqnarray}
\label{eqn:goal}
{\mathbb{M}^\star}={\arg\max_{\mathbb{M}\in\mathcal{S}}\mathrm{Prob}\!\left[\arg\max g_\mathbb{M}\!\left(\mathbf{x}_n;\boldsymbol{\theta}_\mathbb{M}^\star\right)=y_n\mid\left(\mathbf{x}_n,\mathbf{y}_n\right)\in\mathcal{D}_\mathrm{val}\right]},\\
\label{eqn:model_training}
\mathrm{where}\quad{\boldsymbol{\theta}_\mathbb{M}^\star}={\arg\min_{\boldsymbol{\theta}}\mathbb{E}\!\left[\left\|\mathbf{y}_n-g_\mathbb{M}\!\left(\mathbf{x}_n;\boldsymbol{\theta}\right)\right\|^2\mid\left(\mathbf{x}_n,\mathbf{y}_n\right)\in\mathcal{D}_\mathrm{train}\right]}.
\end{eqnarray}
\end{tiny}
Following this setting, the straightforward way is to sample a few architectures, $\mathbb{M}$, from $\mathcal{S}$, train them using Eqn~\eqref{eqn:model_training}, and evaluate them using Eqn~\eqref{eqn:goal} to find the best one. This strategy was widely used by early NAS approaches~\cite{zoph2016neural,real2017large,zoph2018learning}. Albeit stable and effective, it requires an individual training process for each sampled architecture, and thus suffers heavy computational overhead.

To accelerate the search process, researchers noticed that if two architectures are similar, \textit{e.g.}, containing a few common network layers, then a considerable amount of computation for training them can be shared. The so-called one-shot architecture search~\cite{brock2017smash} is a typical example of weight-sharing NAS, in which a \textbf{super-network}, $\mathbb{S}$, is trained, which covers all possible architectures as its \textbf{sub-networks}. For example, if the search process is to construct an architecture with $L$ layers and each layer has a choice among $O$ operators, then the super-network contains $L$ layers but each layer stores the model parameters for all $O$ operators. During the training process, these $O$ operators can either compete with each other in a weighted sum~\cite{liu2018darts}, or be sampled with a probability~\cite{pham2018efficient,chu2019fairnas}. After the super-network is sufficiently optimized, a number of sub-networks can be sampled from it and get evaluated with reduced computational costs. 

\begin{figure*}[ht]
\centering
\includegraphics[width=0.68\linewidth]{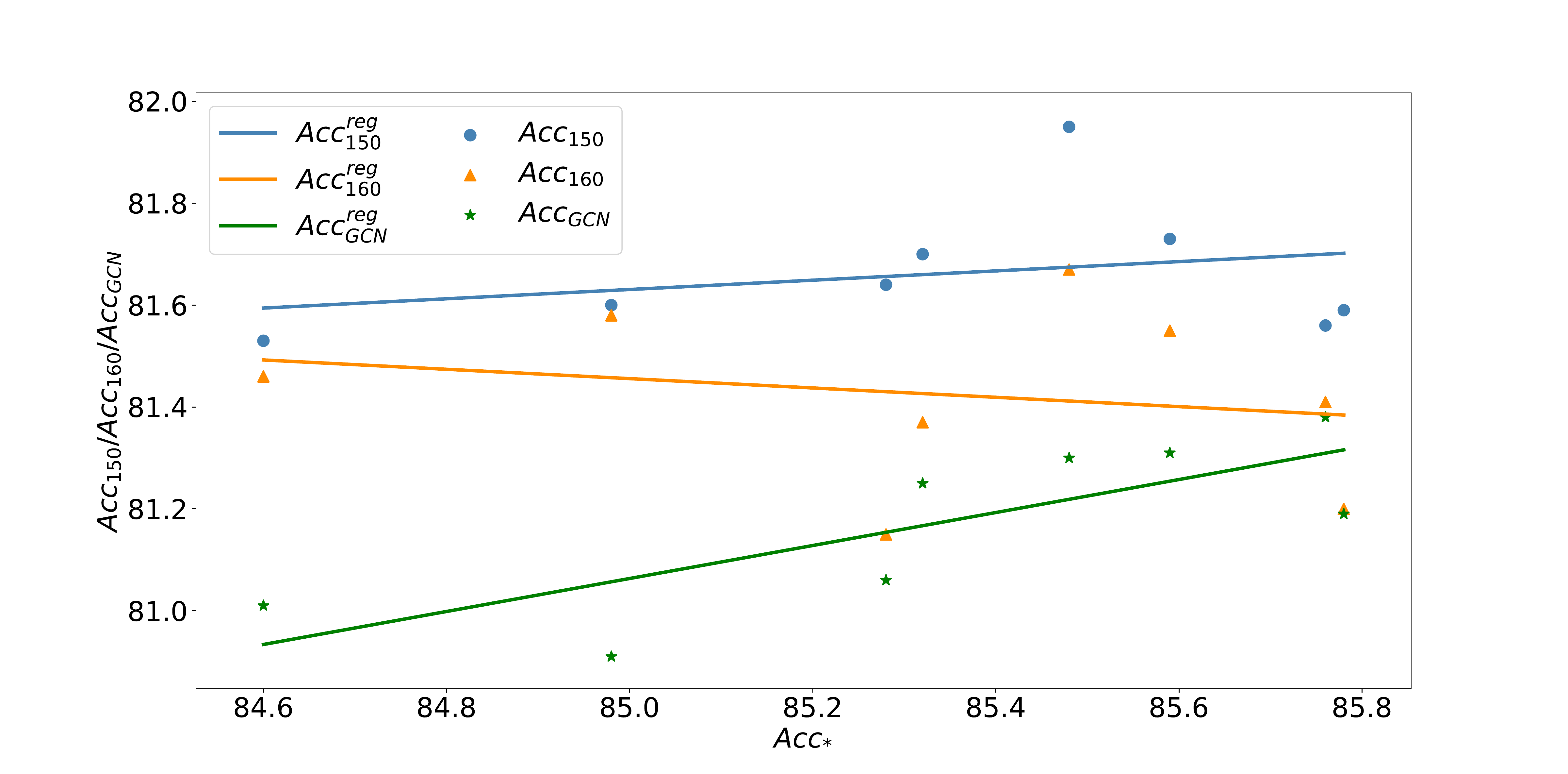}
\caption{Visualization of data in Figure~\ref{fig:inconsistency} and the corresponding linear regression curves. }
\label{fig:appfig1}
\end{figure*}

\section{Graph Construction for GCN}\label{graph_con}

As described in ~\ref{gcn_nas}, we simply define the adjacent matrix, $\mathbf{A}$, to be the similarity matrix. There are typically two ways of defining inter-node similarity. The \textbf{assigned} way assumes that similarity is only determined by the Hamming distance between two sub-networks -- since all edges connect nodes with a Hamming distance of $1$, this is equivalent to setting a fixed weight to each edge, which we use $e^{-0.5}$ throughout this work. Beyond this naive manner, another way named \textbf{measured} similarity can take the changed cell into consideration, for which we make statistics over multiple architectures and calculate the correlation coefficient as similarity. We shall see in experiments that both ways lead to similar accuracy in either GCN ranking or the final architecture.

In this work, we simply follow the Gray coding scheme to encode each sub-network into a fixed-dimensional vector. With Gray code, we guarantee that the Hamming distance between sub-architectures reflects the level of difference in the properties. In our search space that has $19$ cells and $6$ choices for each cell, a $3$-bit binary code is sufficient to encode each cell, and thus the length of each feature vector is $57$. This coding scheme is easy to generalize to more complex larger search spaces.

\section{Dataset, Settings, and Implementation Details}\label{exp_details}

We conduct all experiments on ILSVRC2012~\cite{russakovsky2015imagenet}, a subset of the ImageNet dataset~\cite{deng2009imagenet}. It is a popular image classification benchmark that has $1\rm{,}000$ object categories, $1.28\mathrm{M}$ training images, and $5$K validation images. For the search stage, we randomly sample $100$ from $1\rm{,}000$ classes and split the sampled training images into two parts, with the first part containing $90\%$ samples used for super-network training and the second part with the remaining $10\%$ images used for sub-network evaluation. Unless otherwise specified, when an architecture is evaluated from scratch, we use the standard training set with all $1\rm{,}000$ classes and the input image size is $224\times224$.

Throughout the experiments, we investigate a search space in which all architectures are chain-styled and contain $19$ cells. Each cell is an inverted residual block as described in MobileNet-v2~\cite{sandler2018mobilenetv2}, with a changeable kernel size in $\left\{3,5,7\right\}$ and a changeable expansion ratio in $\left\{3,6\right\}$. Since there are ${2\times3}={6}$ choices for each cell, the overall search space contains $6^{19}$ different architectures. The channel configuration is identical to that of FairNAS~\cite{chu2019fairnas}, so that most architectures in the search space obey the mobile setting, \textit{i.e.}, with no more than $600\mathrm{M}$ FLOPs.

The super-network is trained for $150$ epochs with a batch size of $1\rm{,}024$, distributed over $8$ NVIDIA Tesla-V100 GPUs. The training process takes about $13$ hours. An SGD optimizer with a momentum of $0.9$ and a weight decay of $4\times10^{-5}$ is used. The initial learning rate is set to be $0.18$ and it gradually anneals to $0$ following the cosine schedule.

In the iterative architecture search process, the search space is partitioned into $3$ segments which contain $7$, $6$, and $6$ cells, respectively. With the super-cell composed with the previous searched segment, there are also $7$ cells in the latter two segments. For each segment, $2\rm{,}000$ architectures are sampled, 1\rm{,}800 for GCN training and the rest $200$ for validation. We have also tried a configuration of $M=5\rm{,}000$ for each segment, which will be discussed in the ablation study. 

In each subspace, the GCN is composed of $2$ hidden layers, each having $512$ neurons. An Adam optimizer with an initial learning rate of $0.01$ (decayed by a multiplier of $0.1$ at $1/2$ and $3/4$ of the training process) and a weight decay of $5\times10^{-4}$ is used to tune the parameters. The training process of GCN elapses $600$ epochs, which take around $10$ minutes on a single GPU. After the GCN is well optimized, we sample $100$ top-ranked sub-networks predicted by it and re-evaluate each of them using sub-network sampling.

When training the searched architecture from scratch, an RMS-Prop optimizer with an initial learning rate of $0.128$ (decayed every $3$ epochs by a factor of $0.963$) and a weight decay of $1\times10^{-5}$ is adopted. The batch size is set to be $2\rm{,}048$ (distributed over $8$ GPUs) and learning rate warm-up is applied for the first $5$ epochs. Dropout with a drop rate of $0.2$ is added to the last convolutional layer of the network.

\section{Performance with Hardware Constraints}
\label{hardware}

Here we list performance and details of architectures selected in ~\ref{flex_nas} in Table ~\ref{table:hardware}. 

\begin{table}[h]
\begin{center}
\caption{Performance of discovered architectures under different multi-adds levels}
\label{table:hardware}
\begin{tabular}{lccccc}
\hline
\textbf{\multirow{2}{*}{Architecture}} & \multicolumn{2}{c}{\textbf{Test Acc. (\%)}} & \textbf{Params} & $\times+$ \\
&                            \textbf{top-1} & \textbf{top-5} & \textbf{(M)} & \textbf{(M)} \\
\hline
One-Shot-NAS-GCN-A & 75.6 & 92.7 & 4.6 & 393 \\
One-Shot-NAS-GCN-B & 75.5 & 92.7 & 4.4 & 383 \\
One-Shot-NAS-GCN-C & 75.4 & 92.5 & 3.9 & 360 \\
\hline
\end{tabular}
\end{center}
\end{table}

\end{document}